\documentclass{article}
\usepackage[a4paper, margin=1in]{geometry}
\usepackage{graphicx}
\usepackage{amsfonts}
\usepackage{lscape}
\usepackage{amsthm}
\usepackage{amsmath}
\usepackage{tikz-cd}
\usetikzlibrary{positioning}
\usepackage{amssymb,stmaryrd}
\usepackage{url}
\usepackage{float}
\usepackage{verbatim}
\usepackage{multirow}
\restylefloat{table}
\usepackage{authblk}
\usepackage{subcaption}
\usepackage{mwe}
\usepackage{enumitem}
\usepackage{listofitems}
\usepackage[outline]{contour} 
\contourlength{1.4pt}
\usepackage{multirow}
\usepackage{booktabs} 
\usepackage{hyperref}
\usepackage{array}
\usepackage[ruled,vlined]{algorithm2e}
\usepackage{booktabs}
\usepackage[left]{lineno}
\usepackage{graphicx}
\usepackage{subcaption}
\usepackage{caption}
\usepackage[margin=1in]{geometry}
\usepackage{tabularx} 

\usepackage[sorting=none]{biblatex}
\DeclareUnicodeCharacter{0229}{\c{e}}

\theoremstyle{plain}

\newtheorem*{lemma*}{Lemma}

\theoremstyle{definition}

\numberwithin{theorem}{section}
\numberwithin{definition}{section}
\numberwithin{lemma}{section}
\numberwithin{proposition}{section}
\numberwithin{corollary}{section}
\numberwithin{notation}{section}
\numberwithin{remark}{section}
\numberwithin{example}{section}

\title{Adaptive Nonlinear Vector Autoregression: Robust Forecasting for Noisy Chaotic Time Series}
\author[1]{Sherkhon Azimov\thanks{Corresponding author. Email: sherxonazimov94@pusan.ac.kr}}
\author[1,2,3]{Susana L\'opez-Moreno}
\author[2,4]{Eric Dolores-Cuenca}
\author[5]{Sieun Lee}
\author[6]{Jae-Il Kwon}
\author[1,7]{Sangil Kim\thanks{Corresponding author. Email: sangil.kim@pusan.ac.kr}
}

\affil[1]{Department of Mathematics, Pusan National University, Republic of Korea}
\affil[2]{Industrial Mathematics Center, Pusan National University, Republic of Korea}
\affil[3]{Humanoid Olfactory Display Center, Pusan National University, Republic of Korea}
\affil[4]{Department of Mathematics, Yonsei University, Republic of Korea}
\affil[5]{Innovation Center for MathScience Research\&Education, Pusan National University, Republic of Korea}
\affil[6]{Marine Natural Disaster Research Department, Korea Institute of Ocean Science and Technology, Republic of Korea}
\affil[7]{Institute for Future Earth, Pusan National University, Republic of Korea}

\date{\today}
\usepackage{tocloft}

\begin{document}

\maketitle
\begin{abstract}
Nonlinear vector autoregression (NVAR) and reservoir computing (RC) have shown promise in forecasting chaotic
dynamical systems, such as the Lorenz-63 model and El Niño–Southern Oscillation. However, their reliance on fixed nonlinear transformations—polynomial expansions in NVAR or random feature maps in RC—limits their adaptability to high noise or complex real-world data. Furthermore, these methods also exhibit poor scalability in high-dimensional settings due to costly matrix inversion during optimization. We propose a data-adaptive NVAR model that combines delay-embedded linear inputs with features generated by a shallow, trainable multilayer perceptron (MLP). Unlike standard NVAR and RC models, the MLP and linear readout are jointly trained using gradient-based optimization, enabling the model to learn data-driven nonlinearities, while preserving a simple readout structure and improving scalability. Initial experiments across multiple chaotic systems, tested under noise-free and synthetically noisy conditions, showed that the adaptive model outperformed in predictive accuracy the standard NVAR, a leaky echo state network (ESN)—the most common RC model— and a hybrid ESN, thereby showing robust forecasting under noisy conditions.
\end{abstract}


\section{Introduction}\label{Sec:intr}
\subsection{Motivation}
Time series analysis and forecasting have become essential tools in many fields, such as climate science \cite{bochenek2022machine}, \cite{forster2024indicators}, \cite{hansen2010global}, \cite{wilson2016last},  finance \cite{dingli2017financial}, \cite{hamilton2020time}, \cite{takahashi2019modeling}, healthcare and medicine \cite{esteban2016predicting}, \cite{lipton2015learning},  \cite{rajkomar2018scalable} and transportation \cite{lv2014traffic}. Numerous cutting-edge methods have been proposed to address the difficulties in modeling complex temporal dependencies in these domains. 

Reservoir computing (RC) is a machine learning paradigm introduced in the early 2000s that uses reservoirs to learn spatiotemporal features in time-series data 
\cite{jaeger2004harnessing}, \cite{maass2002real}, \cite{reservoir_computing}. Optimized RC frameworks can address highly challenging tasks, such as chaotic or complex spatiotemporal behaviors \cite{pathak2018model}, \cite{pathak2017using}. In echo state networks (ESN)\cite{RC}, the most commonly used RC model, the reservoir consists of a recurrent neural network \cite{recNN}, where the weights are randomly sampled and fixed (i.e., untrained). By fixing the reservoir weights, training is reduced to a simple least squares estimation problem, rather than the costly, fully nonlinear optimization required for standard recurrent neural networks. Additionally, a universal approximation theorem was established in \cite{gonon2019reservoir} demonstrating that reservoir computers can approximate the causal, time-invariant functionals of stationary stochastic processes. Although RC offers the fast computation and lightweight design, it presents several challenges, as described in \cite{challenges_RC}. 

Next-generation reservoir computing (NG-RC) was introduced in \cite{NG-RC} to address three major limitations of conventional reservoir computing: (i) the lack of interpretability caused by the reservoir functioning as a black box, (ii) the substantial memory demands imposed by the randomly sampled matrices, and (iii) the presence of numerous hyperparameters requiring tuning— such as the spectral radius, input scaling and leaking rate. NG-RC is a formalization of the nonlinear vector autoregression (NVAR) framework. It was also shown in \cite{explaining_RC} that RC is mathematically equivalent to NG-RC (or NVAR) when the activation function of the reservoir is set to be the identity function.

A comparative study \cite{Comparative_study} showed that NVAR outperforms long short-term memory networks, gated recurrent units, and several ESN architectures in the prediction of chaotic time series. In \cite{NG-RC}, it was demonstrated that NVAR performs well on three challenging RC benchmark problems. However, few studies have conducted experiments under noisy conditions, such as \cite{SHENG2012186} for an ESN that was based on dual estimation or \cite{doan2021short} for a hybrid physics informed ESN.

\begin{figure}[h!]
\centering
\includegraphics[width=1.0\linewidth]{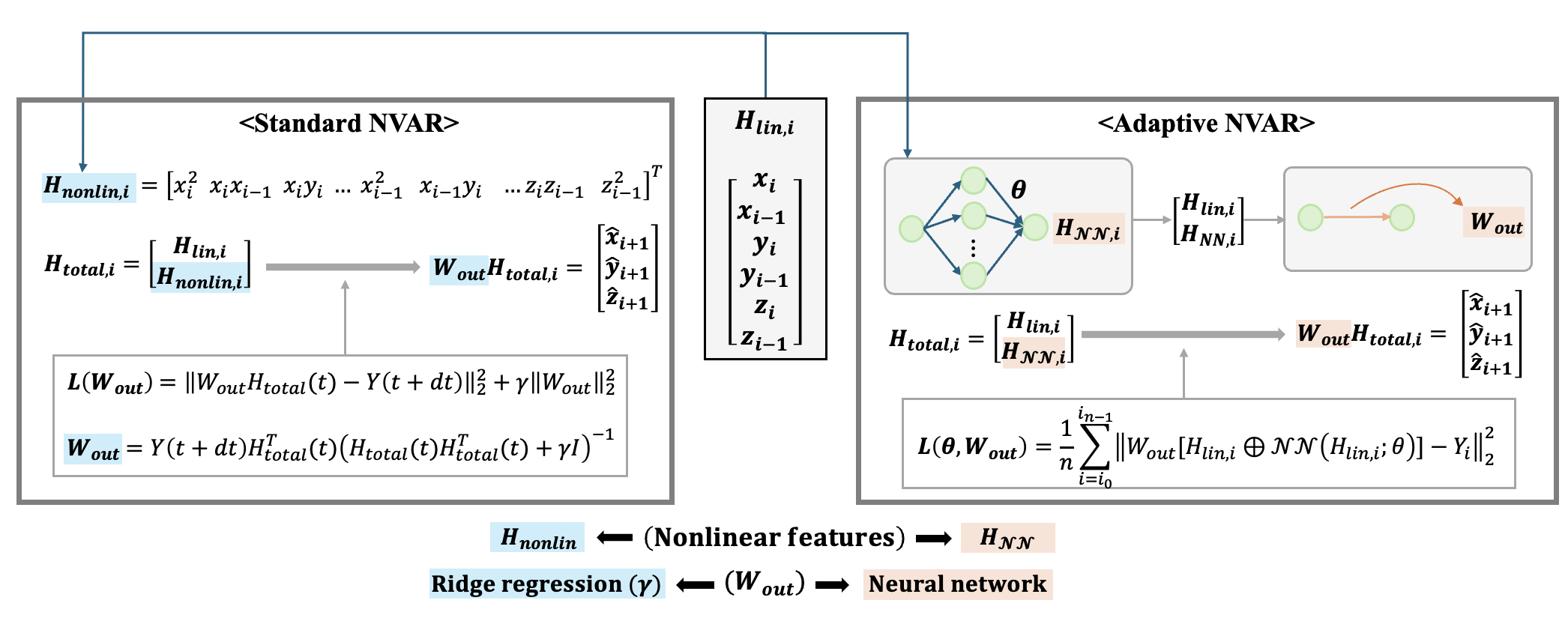}
\caption{Comparison of the standard NVAR (left) and Adaptive NVAR (right). In the standard formulation, the nonlinear feature vector is created as a (quadratic) polynomial and the linear readout matrix $W_{\text{out}}$ is computed as the closed-form solution of a least-squares regression with Tikhonov regularization (ridge regression). In contrast, the adaptive model employs a trained MLP to generate $H_{\mathcal{NN}}$, while $W_{\text{out}}$ is treated as a trainable weight matrix within the skip-connection architecture of the adaptive model via gradient descent.}  
\label{fig:hybrid_nvar}
\end{figure}

Standard NVAR uses fixed polynomial basis functions to incorporate nonlinearity, thereby limiting its adaptability in handling high-dimensional or noisy data. Moreover, the linear readout matrix in both RC and NVAR depends on the nonlinear feature vector and is therefore susceptible to accumulating errors originating from this vector. This linear readout matrix computation, which we introduce in Section~\ref{Sec:RC}, also requires a matrix inversion, which can become a significant computational bottleneck for large feature vectors.
A kernel ridge regression approach for the NVAR with polynomial basis functions was introduced in~\cite{infinitedimensional}. That study also explored replacing the polynomial kernel by a Volterra kernel. However, kernel methods suffer from scalability issues when applied to large-scale data.

We propose a data-adaptive NVAR model, termed Adaptive NVAR, that addresses the limitations of standard NVAR, by replacing the quadratic nonlinear feature vector with a learnable, shallow multilayer perceptron (MLP). This MLP transforms the delay-embedded inputs in a data-adaptive manner and addresses the explosion of the dimensionality of the feature vector of standard NVAR for high dimensional models. 
By using an MLP, we trade the interpretability offered by the NVAR technique for scalability and improved performance on noisy data in chaotic systems, an essential trade-off for forecasting real-world phenomena. 
The scalability of reservoir computing models has been addressed previously in papers such as \cite{baur2021predicting}, where the authors employ generalized local states, a generalization of the reconstructed local states introduced in \cite{parlitz2000prediction}, which were then used for the prediction of large chaotic systems using reservoir computing in \cite{pathak2018model}. This technique addresses heterogeneity and high dimensionality by partitioning variables and defining local neighborhoods via correlation-based distances, which changes the state construction pipeline. Although successfully employed for RC models, incorporating local states mechanisms would alter the NVAR feature construction and therefore yield a non-NVAR architecture. Adaptive NVAR instead targets NVAR’s scalability problems and the rigidity of the fixed polynomial feature map under noise, by learning the nonlinear feature transformation via a shallow trainable network.

We highlight the difference between standard NVAR and Adaptive NVAR in Figure~\ref{fig:hybrid_nvar}. A key distinction in the design of our method is the joint optimization of both the nonlinear feature vector and the linear readout matrix through a gradient-based method, within a skip-connection architecture~\cite{wienercybernetics}.
This end-to-end training enables the linear readout matrix and nonlinear features to be learned simultaneously, yielding an optimal joint configuration and decoupling the readout from any fixed feature representation. As a result, the learned features are explicitly tuned to the data, and the model's adaptability to chaotic and noise-perturbed systems is enhanced, making the approach well suited for forecasting geophysical processes such as ocean salinity, sea surface temperature and climate indices.

The work in~\cite{hitandrun} is closely related to our approach, as it also employs skip-connection layers and neural networks to replace the nonlinear feature vector of a standard NVAR. However, the architecture described in that work uses fixed weights, and the readout matrix is still computed through least-squares regression, which limits their application to noise-free data.

In this paper, we conduct experiments on the Lorenz-63 chaotic system, Mackey-Glass system and Lorenz-96 system with 100 variables, starting with noise-free data and subsequently introducing varying levels of synthetic noise. We consider dynamical systems with known equations to allow for benchmarking and for experiments to be performed under controlled amounts of noise. Real-world data is inherently noisy and partially observable, so we leave the application to real datasets as future work. We compare the forecasting performance of Adaptive NVAR with the standard NVAR, a leaky ESN and a hybrid ESN \cite{pathak2018hybrid}, which have been identified earlier in \cite{Comparative_study} as the three best-performing models under noise-free data. 

This paper is organized as follows. Section~\ref{Sec:Background} provides preliminary concepts on RC and the standard NVAR, and Section~\ref{Sec:Hybrid_NVAR} details the formulation and algorithm of Adaptive NVAR. In Section~\ref{Sec:res}, we showcase the results of our experiments. Section~\ref{Sec:discussion} discusses these experimental results, the scalability of the proposed method, as well as its limitations and directions for future work.

\subsection{Background}\label{Sec:Background}

This section provides a summary of the concept of RC and NVAR. The formulation of the hybrid ESN (HESN), employed for comparative purposes in our experiments, is not presented here. It is detailed in \cite{pathak2018hybrid}, but its central idea consists of the addition to Equation~\ref{eq:RC} of an imperfect mathematical knowledge-based term $(1 + \epsilon) b$, where the error parameter $\epsilon$ is set to 0.1.   

\subsubsection{Reservoir Computing}\label{Sec:RC}
We introduce traditional RC, extended to include leaky integrator neurons. 
Let $X_i=[x_{1,i},x_{2,i},\dots,x_{d,i}]\in\mathbb{R}^{d}$ be a time-series training data for time $i\in\{1,\dots,T\}$. This linear vector is introduced into the reservoir through the input layer with a fixed, randomly sampled matrix $W_{\text{in}}$. The dynamic reservoir system is governed by the following iterative update
\begin{equation}\label{eq:RC}
\begin{aligned}
r_{i+1}&=(1-\alpha)r_i +\alpha f(Ar_i+W_{\text{in}}X_i+b),\\
\hat{Y}_{i+1}&=W_{\text{out}}r_{i+1},
\end{aligned}
\end{equation}
where $r_i$ is an $m$-dimensional reservoir state vector, typically with $m>d$. The parameter $\alpha\in [0,1]$ denotes the leaking (or decay) rate of the nodes, $f$ is an activation function (commonly, $f(x)=tanh(x)$), $A$ is a fixed, randomly sampled connectivity matrix, and $b$ is a bias vector. Finally, $W_{\text{out}}$ is trained on the data and obtained by least-squares regression with Tikhonov regularization. The loss function is defined as
\begin{equation*}\label{eq:loss_NVAR}
L(W_{\text{out}})=\|W_{\text{out}}H_{\text{total}}(t)-Y(t+dt) \|_2^2+\gamma \|W_{\text{out}} \|_2^2,
\end{equation*}
where $H_{\text{total}}$ denotes the block of data generated by the training points and $Y(t+dt)$ is the target output. The closed-form solution of this equation \cite{tikhonov1977solutions} is given by 
\begin{equation}\label{eq:Wout}
W_{\text{out}}=Y H_{\text{total}}^\intercal(H_{\text{total}}H_{\text{total}}^\intercal+\gamma I)^{-1},
\end{equation}
where $I$ is the identity matrix. After computing the linear readout weights, forecasts are generated iteratively with each output fed back as an input for the next prediction step.

\subsubsection{Nonlinear Vector Autoregression}\label{Sec:NVAR}
Let $X_i=[x_{1,i},x_{2,i},\dots,x_{d,i}]\in\mathbb{R}^{d}$ be the $d$-dimensional time-series training data for $i\in\{1,\dots,T\}$. The standard NVAR framework—specifically the NG-RC formulation in \cite{NG-RC}—constructs the feature vector as follows
\begin{equation*}
H_{\text{total}}=b\oplus H_{\text{lin}}\oplus H_{\text{nonlin}},
\end{equation*}
where $b$ is a bias constant and $\oplus$ denotes concatenation of vectors. 

The linear feature vector $H_{\text{lin}}$ at step $i$ \cite{kantz2003nonlinear} is defined as
\begin{equation*}
\begin{aligned}
H_{\text{lin},i}&=X_i\oplus X_{i-s}\oplus X_{i-2s} \oplus\cdots\oplus X_{i-(k-1)s}\\
&=\begin{bmatrix}
X_i\\
X_{i-s} \\
X_{i-2s} \\
\vdots \\
X_{i-(k-1)s}
\end{bmatrix},
\end{aligned}
\end{equation*}
where $k$ is the delay parameter, meaning the vector includes $k-1$ previous time steps, each spaced by $s$, the number of skipped steps between consecutive observations. Thus, $H_{\text{lin},i}$ contains $d_{\text{lin},i}=d\times k$ components in total.

This formulation implies that, while traditional RCs require long warm-up periods—in the order of thousands of data points \cite{NG-RC}—to ensure that the reservoir state does not depend on initial conditions, the standard NVAR requires only $k\times s$ steps to construct the linear feature vector. For simplicity, we assume $s=1$ for both the standard and Adaptive NVAR, consistent with the formulation in Section~\ref{Sec:Hybrid_NVAR} and the grid search hyperparameter range adopted from \cite{Comparative_study}.
It allows for a direct comparison between both NVAR methods without adding extra degrees of freedom.

In contrast, the nonlinear feature vector $H_{\text{nonlin}}$ is derived directly from the linear feature vector as a nonlinear function of $H_{\text{lin}}$. In \cite{NG-RC}, this nonlinear function is defined as a quadratic polynomial, and $H_{\text{nonlin},i}$ is composed of the $m:=d_{\text{nonlin},i}=\frac{(dk)(dk+1)}{2}$ unique monomials of the outer product $H_{\text{lin},i}\otimes H_{\text{lin},i}$. The loss function is then defined as
\begin{equation*}
L(W_{\text{out}})=\|W_{\text{out}}H_{\text{total}}(t)-Y(t+dt) \|_2^2+\gamma \|W_{\text{out}} \|_2^2,
\end{equation*}
which is identical for RC and NVAR. As before, $W_{\text{out}}$ is obtained via least-squares regression with Tikhonov regularization, which has the closed-form solution
\begin{equation*}
W_{\text{out}}=Y H_{\text{total}}^\intercal(H_{\text{total}}H_{\text{total}}^\intercal+\gamma I)^{-1}.
\end{equation*}

In \cite{NG-RC}, the target output $Y$ is defined as the difference $X_{i+1}-X_i$, and the linear readout then takes an Euler-like form
\begin{equation}\label{eq:diff}
\hat{X}_{i+1}=\hat{X}_i+W_{\text{out}}H_{\text{total}}.
\end{equation}
We also adopt this difference-based training in the Adaptive NVAR model.

One of the challenges in standard NVAR is the selection of appropriate values for $\gamma$ and $k$, as the model is highly sensitive to these parameters. Table~\ref{tab:hyperparams} and Supplementary Tables 3–6 document our parameter optimization process to ensure a fair comparison between the standard and Adaptive NVAR in the experiments presented in Section~\ref{Sec:res}.

\section{Adaptive Nonlinear Vector Autoregression}\label{Sec:Hybrid_NVAR}
In this section, we present the data-adaptive NVAR model. Figure~\ref{fig:hybrid_nvar} contrasts the conventional NG-RC (or standard NVAR) formulation with the proposed Adaptive NVAR. Both approaches have an identical construction of the linear feature vector and computation of the output vector. The distinction arises in the treatment of the nonlinear feature vector: while standard NVAR constructs a fixed nonlinear vector derived from the linear features, Adaptive NVAR employs a shallow MLP to learn this representation. Furthermore, we integrate $W_{\text{out}}$ into the training architecture via a skip connection, enabling its joint optimization with the nonlinear feature vector. This eliminates the need for matrix inversion in \eqref{eq:Wout}, replacing it with gradient-based optimization.

Let $X_i=[x_{1,i},x_{2,i},\dots,x_{d,i}]\in\mathbb{R}^{d}$ be a vector in a time series with time index $i\in\{1,2,\dots,T\}$, where $T$ is the total number of times steps. Following the standard NVAR framework, we construct a delay-embedded input vector as follows
\begin{equation*}
H_{\text{lin},i}=X_i\oplus X_{i-s}\oplus X_{i-2s}\oplus\cdots\oplus X_{i-(k-1)s}\in\mathbb{R}^{dk},
\end{equation*}
where $k$ denotes the number of time delays, $\oplus$ indicates vector concatenation and $s$ specifies the number of time steps skipped between each delay. For simplicity we assume $s=1$. This formulation ensures the model initialization period coincides with that of the conventional NVAR approach.

The ability of neural networks, particularly MLPs, to approximate complex nonlinear functions is well established. As demonstrated in \cite{hornik1989multilayer}, MLPs serve as universal function approximators that, with sufficient representational capacity, can represent a wide class of nonlinear mappings. Owing to their ability to capture nonlinear temporal dependencies, neural networks have also been extensively used in time series forecasting \cite{rao2024predictingchaoticbehaviorusing}, \cite{weigend2018time}. For a rigorous mathematical treatment of neural network architectures and deep learning, we refer readers to an earlier study \cite{higham2019deep}, which provides a comprehensive introduction tailored to applied mathematicians. Building upon these foundations, we define an MLP denoted as $\mathcal{NN}(\cdot;\theta)$, which transforms the delay embedding vector into the following nonlinear feature representation

\begin{equation}\label{eq:NN}
H_{\mathcal{NN},i}=\mathcal{NN}(H_{\text{lin},i};\theta)\in\mathbb{R}^{d_{\mathcal{NN},i}},
\end{equation}
where $\theta$ denotes the trainable parameters of the network and $d_{\mathcal{NN},i}$ specifies the output dimensionality of the nonlinear feature space produced by the neural network. To match the NVAR approach, we set $d_{\mathcal{NN},i}=\frac{(dk)(dk+1)}{2}$ for systems of low to moderate dimensionality, and $d_{\mathcal{NN},i}=c\,dk$ for high-dimensional systems, where $c\in\mathbb{N}$ is a tunable constant. We then compute $H_{\mathcal{NN},i}$ as

\begin{equation}\label{eq:H_NN}
H_{\mathcal{NN},i} = W \bigl( \tanh( W_{\text{in}} H_{\text{lin},i} +b_1) \bigr)+b_2,
\end{equation}
where $W$ and $W_{\text{in}}$ are trainable weight matrices and $b_2$ is a bias vector. This neural transformation generalizes the fixed nonlinear feature map traditionally used in NVAR. Finally, the total feature vector is constructed by concatenating the linear and nonlinear components,
\begin{equation}\label{eq:concatenation}
H_{\text{total},i}=H_{\text{lin},i}\oplus H_{\mathcal{NN},i}\in\mathbb{R}^{dk+d_{\mathcal{NN},i}}.
\end{equation}
Relative to the standard NVAR, this formulation does not include a bias constant.

For prediction, we employ a linear readout defined as
\begin{equation*}\label{eq:linear_readout}
\hat{Y}_{i}=\hat{X}_{i+1}-\hat{X}_{i}=W_{\text{out}}H_{\text{total},i},
\end{equation*}
since, as in Equation~\eqref{eq:diff}, the model forecasts the difference between states $\hat{X}_{i+1}-\hat{X}_i$, rather than the absolute state. Here, the weight matrix $W_{\text{out}}\in\mathbb{R}^{d\times (dk+d_{\mathcal{NN},i})}$ is trainable. To further clarify this skip-connection architecture, the corresponding pseudocode is presented in Algorithm~\ref{algorithm:adaptive_NVAR}.\bigskip

\begin{algorithm}[H]
\caption{Adaptive NVAR Model}
\label{algorithm:adaptive_NVAR}
\KwIn{Linear features \texttt{H\_lin} in $ \mathbb{R}^{dk}$}
\KwOut{Predicted value in $\mathbb{R}^{d}$}

\textbf{MLP}\\
\hspace*{1em}\texttt{Linear(input\_dimension=dk, hidden\_dimension)} \\
\hspace*{1em}\texttt{Tanh()} \\
\hspace*{1em}\texttt{Linear(hidden\_dimension, output\_dimension=d\_NN)}

\vspace{0.5em}
\textbf{Readout} \\
\hspace*{1em}\texttt{Linear(dk + d\_NN, d, bias=False)}

\vspace{0.5em}
\textbf{Forward pass:} \\
\hspace*{1em}\texttt{H\_NN = MLP(H\_lin)} \\
\hspace*{1em}\texttt{H\_total = concatenation([H\_lin, H\_NN], dim=1)} \\
\hspace*{1em}\texttt{return Readout(H\_total)}

\end{algorithm}

\medskip
Adaptive NVAR is trained in end-to-end manner, where both the nonlinear neural feature parameters $\theta$ and the linear readout weights $W_{\text{out}}$ are jointly optimized by minimizing the prediction loss

\begin{equation}
\begin{aligned}
L(\theta,W_{\text{out}})&=\frac{1}{n}\sum\limits_{i=i_0}^{i_{n-1}} \|\hat{Y}_i-Y_i\|_2^2 \\
&=\frac{1}{n} \sum\limits_{i=i_0}^{i_{n-1}}\|W_{\text{out}} H_{\text{total},i}-Y_i\|_2^2 \\
&=\frac{1}{n} \sum\limits_{i=i_0}^{i_{n-1}}\|W_{\text{out}} [H_{\text{lin},i}\oplus\mathcal{NN}(H_{\text{lin},i};\theta)]-Y_i\|_2^2, 
\end{aligned}
\end{equation}
where $n$ denotes the number of training points and $i_0$ is the first valid training index. Supplementary Example 1 illustrates a simple working case that clarifies the indexing and the formulation of the model.

In contrast to conventional NVAR approaches, which solve the linear readout analytically, our adaptive approach treats all components as differentiable and trainable, allowing full backpropagation through both the delay embedding and nonlinear transformation.
Model parameters are optimized using the AdamW optimizer \cite{kingma2014adam}, together with a ReduceLROnPlateau learning-rate scheduler, gradient clipping, and early stopping to improve convergence and training stability.

Unlike RC, which is mathematically equivalent to NVAR under certain conditions~\cite{explaining_RC}, the proposed Adaptive NVAR method is not mathematically equivalent to an NVAR model even when the MLP uses the identity function as its activation. In fact, replacing tanh with the identity function renders the proposed method linear.

To understand the relationship between the architecture of Adaptive NVAR and standard NVAR, let us denote the span of the shallow MLP in Adaptive NVAR by  \[\sum_n=\operatorname{span}\{\tanh(W\cdot x+b_1)\ | \ W\in \mathbb{R}^n, \ b_1\in \mathbb{R}\}.\]
From the work of~\cite{Multilayerapprox}, if $\mu$ is a non-negative finite measure on $\mathbb{R}^n$, with compact support and absolutely continuous with respect to the Lebesgue measure, then $\displaystyle\sum_n$ is dense in $L^p(\mu)$, for $ 1\leq p< \infty$. Consequently, a shallow MLP can approximate the vector of polynomial features used in any NVAR architecture. Therefore, Adaptive NVAR can approximate any NVAR architecture, including but not limited to NG-RC, which specifically uses quadratic polynomials. As a result, any dynamical system that can be approximated by standard NVAR can also be approximated by Adaptive NVAR.

During early experiments, we tested MLPs with different numbers of layers and hidden units. We found that a shallow MLP could effectively extract important nonlinear features from the linear delay-embedded inputs. 
However, the proposed framework is not limited to this choice, and one could include other architectures if needed for specific applications.

\section{Results}\label{Sec:res}
This section presents the experimental results comparing the proposed Adaptive NVAR model with three established reservoir computing (RC) benchmarks: standard NVAR, ESN, and HESN. These models were identified as the best-performing RC architectures for chaotic systems in \cite{Comparative_study}, outperforming other recurrent neural network architectures such as LSTM and GRU.

We evaluate the forecasting performance of the models on three chaotic systems: the one-dimensional Mackey--Glass system, the low-dimensional Lorenz--63 system, and the high-dimensional Lorenz--96 system, with the latter serving to assess scalability. Motivated by findings in a previous study~\cite{revising}, which showed that a widely used graph neural network benchmark contained incorrect values that subsequently affected later research, we designed our benchmark datasets with particular emphasis on correctness and reproducibility. To promote transparency and strengthen the reliability of future studies, we encourage independent replication of all experiments.

All models were optimized in preliminary experiments using the Tree-structured Parzen Estimator (TPE) \cite{akiba2019optuna, bergstra2011algorithms} to ensure a fair comparison. Hyperparameters were selected using a validation set only, while the test set remained completely unseen during optimization to prevent data leakage. Owing to the relatively small search spaces of the standard and Adaptive NVAR models, 20 Optuna trials were performed for each. In contrast, ESN and HESN were optimized using 500 trials because of their substantially larger hyperparameter search spaces. The hyperparameter search spaces considered for each method are summarized in Table~\ref{tab:hyperparams} and were chosen based on commonly used ranges in the reservoir computing literature, particularly the comparative study in \cite{Comparative_study}. The fixed hyperparameters for each model are provided in Supplementary Table~3. To ensure reproducibility, fixed random seeds were used for both dataset generation and the independent model runs.

To evaluate the proposed model under controlled conditions, we used synthetic data generated from known chaotic systems rather than real-world engineering datasets. Observation noise was added to the entire dataset before normalization according to
\begin{equation}
X_{\mathrm{noisy}} =
X_{\mathrm{clean}}
+\sigma\epsilon,\qquad
\epsilon\sim\mathcal{N}\!\left(0,\operatorname{std}(X_{\mathrm{clean}})\right),
\label{eq:noise}
\end{equation}
where $\sigma$ controls the noise level. Four noise levels were considered: $\sigma=0$ (noise-free), $\sigma=0.10$ (low), $\sigma=0.20$ (moderate), and $\sigma=0.30$ (high). For each noise level, three independent noise realizations were generated using fixed random seeds. These realizations were treated as three independent benchmark datasets (Dataset~1--3) to evaluate the robustness of the models under different observation noise realizations and to reduce the influence of any single realization on the reported results.

Model performance was evaluated over multiple overlapping forecast windows extracted from the test trajectory with a stride of 10 time steps. The RMSE was computed for each window and then averaged across all windows. All models were evaluated in a recursive (closed-loop) forecasting setting, where each predicted output was fed back as the input for the subsequent prediction. Forecasts were evaluated at four prediction horizons of 25, 50, 75, and 100 time steps. For all datasets, the time series were intentionally downsampled, as described in detail for each benchmark system below. This downsampling produces sparser temporal observations, thereby increasing the difficulty of the forecasting task for all models. Consequently, prediction errors become saturated beyond approximately 100 prediction steps, making further comparisons uninformative. Therefore, all comparisons are restricted to prediction horizons of up to 100 steps, corresponding to substantially longer physical-time forecasts in the original systems. For all trainable models, the reported RMSE values are the mean and standard deviation over 10 independent runs with different random initializations, whereas the deterministic standard NVAR model was evaluated only once.
\begin{table}[h!]
\centering
\caption{TPE hyperparameter search spaces for each model.}
\label{tab:hyperparams}
\begin{tabularx}{\linewidth}{l l X}
\toprule
\textbf{Method} & \textbf{Parameter} & \textbf{Search space} \\
\midrule
Standard NVAR
& Delay parameter ($k$)
& $\{2,5,10,20,30,40,50\}$ \\
& Ridge regularization ($\lambda$)
& $\{10^{-7},10^{-6},10^{-5},10^{-4},10^{-3},10^{-2},10^{-1}\}$ \\
\midrule

Adaptive NVAR
& Delay parameter ($k$)
& $\{2,5,10,20,30,40,50\}$ \\
& Hidden dimension ($H$)
& $\{32,64,128, 1024, 4096\}$ \\
& Adam learning rate ($\eta$)
& $\{10^{-4},\,5\times10^{-4},\,10^{-3}\}$ \\
\midrule

ESN
& Reservoir size ($N_r$)
& $\{64,128,256,512\}$ \\
& Input scaling ($\sigma_{\mathrm{in}}$)
& $[10^{-2},\,1]$  \\
& Spectral radius ($\rho$)
& $[0.1,\,1.5]$ \\
& Leak rate ($\alpha$)
& $[0.1,\,1.0]$ \\
& Connectivity ($c$)
& $[0.01,\,0.3]$ \\
& Ridge regularization ($\lambda$)
& $\{10^{-7},10^{-6},10^{-5},10^{-4},10^{-3},10^{-2},10^{-1}\}$ \\
\midrule

HESN
& Reservoir size ($N_r$)
& $\{64,128,256,512\}$ \\
& Input scaling ($\sigma_{\mathrm{in}}$)
& $[10^{-2},\,1]$  \\
& Knowledge-based scaling ($\sigma_{\mathrm{kb}}$)
& $[10^{-2},\,1]$  \\
& Spectral radius ($\rho$)
& $[0.1,\,1.5]$ \\
& Leak rate ($\alpha$)
& $[0.1,\,1.0]$ \\
& Connectivity ($c$)
& $[0.01,\,0.3]$ \\
& Ridge regularization ($\lambda$)
& $\{10^{-7},10^{-6},10^{-5},10^{-4},10^{-3},10^{-2},10^{-1}\}$ \\
\bottomrule
\end{tabularx}
\end{table}

\subsection{One-Dimensional System: Mackey–Glass}\label{Sec:MG}

The first benchmark considered is the Mackey--Glass 
delay differential system \cite{mackey1977oscillation}, given by the equation
\begin{equation}\label{eq:MG}
\frac{dx(t)}{dt} = \frac{a \, x(t-\tau)}{1 + \big(x(t-\tau)\big)^c} - b x(t),
\end{equation}
with parameters $a = 0.2,\ b = 0.1$, and $c = 10.$
For $\tau \geq 17$, the system exhibits chaotic dynamics. To generate a time series of $10{,}000$ points, we set $\tau = 17$, integrating the system using a fourth-order Runge–Kutta method with a time step of $\Delta t = 0.1$, followed by downsampling the result by a factor of 10.

An $80/10/10$ partition was employed for warm-up $+$ training/validation, and testing. For all models, the initial 500 points were discarded as a warm-up (washout). The subsequent points were allocated for training, followed by $1{,}000$ points for validation and the final $1{,}000$ points for testing.

Unlike the Lorenz–63 and Lorenz–96 datasets, the Mackey–Glass series was not normalized, since its amplitude exhibits smooth temporal variations without abrupt scale changes and remains bounded within a narrow interval ($x \in [0, 1.5]$, approximately). 

\begin{figure}[h!]
    \centering
    \includegraphics[width=\linewidth]{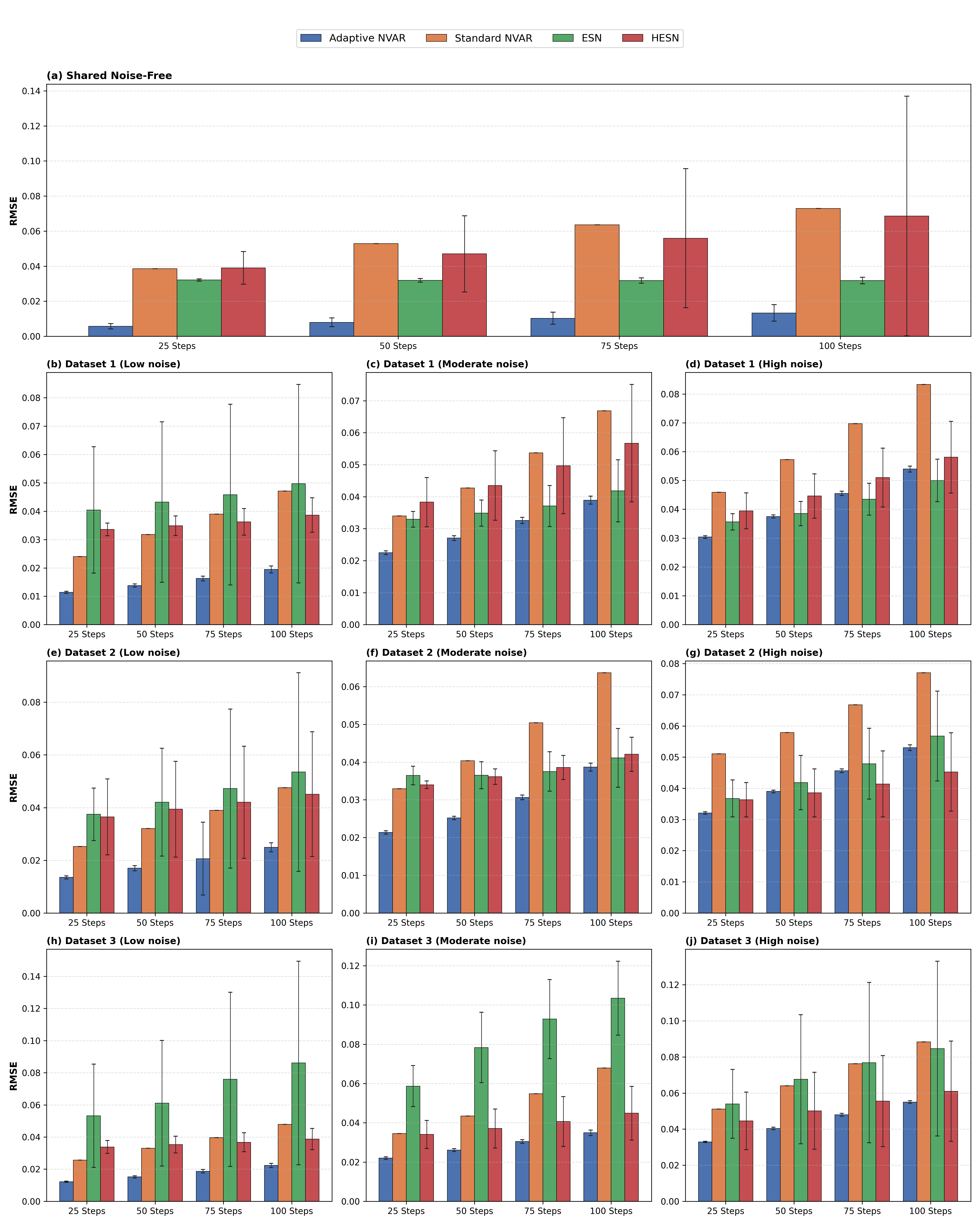}
    \caption{\textbf{Predictive performance and robustness to noise for the Mackey--Glass system.} 
     (a) Noise-free forecasting performance. (b--j) Forecasting performance under three independent noise realizations (Datasets~1--3) at low ($\sigma=0.10$), moderate ($\sigma=0.20$), and high ($\sigma=0.30$) noise levels. Bar heights denote the mean RMSE over overlapping forecast windows, while error bars indicate the standard deviation. Forecasting performance is evaluated at prediction horizons of 25, 50, 75, and 100 time steps}
    \label{fig:mg_results}
\end{figure}

Figure~\ref{fig:mg_results} presents the detailed forecasting performance of the proposed Adaptive NVAR model and the three benchmark models under noise-free and noisy scenarios. The noise-free results are shown in Fig.~\ref{fig:mg_results}(a), while Figs.~\ref{fig:mg_results}(b--j) report the results for three independent noise realizations (Datasets 1--3) under low ($\sigma=0.10$), moderate ($\sigma=0.20$), and high ($\sigma=0.30$) noise. Error bars represent the standard deviation over independent runs.

Under noise-free conditions, Adaptive NVAR consistently achieves the lowest RMSE across all prediction horizons, substantially outperforming standard NVAR, ESN, and HESN. At the longest prediction horizon of 100 steps, Adaptive NVAR attains an RMSE of 0.0133, compared with 0.0318 for ESN, 0.0686 for HESN, and 0.0729 for standard NVAR. These results demonstrate that learning an adaptive nonlinear feature representation significantly improves long-term forecasting accuracy over both the fixed polynomial representation of standard NVAR and the reservoir-based architectures.

The robustness of the models under noise cases is illustrated in Fig.~\ref{fig:mg_results}(b--j). As expected, the prediction error increases with both the forecasting horizon and the observation noise level for all models. Nevertheless, Adaptive NVAR consistently achieves the lowest RMSE across all three independent noise realizations, all noise levels, and all prediction horizons. Moreover, the increase in prediction error is gradual and highly consistent across the three noisy datasets, indicating that the proposed model is considerably less sensitive to individual noise realizations than the competing models.

In contrast, ESN and HESN exhibit noticeably larger variability across both independent runs and different noise realizations, as reflected by their larger error bars. While HESN generally provides improved robustness to observation noise compared with the standard ESN, neither reservoir-based approach matches the forecasting accuracy or stability of Adaptive NVAR. The standard NVAR shows no run-to-run variation but experiences the largest degradation in forecasting accuracy as the prediction horizon increases.

\begin{figure}[h!]
    \centering
    \includegraphics[width=\linewidth]{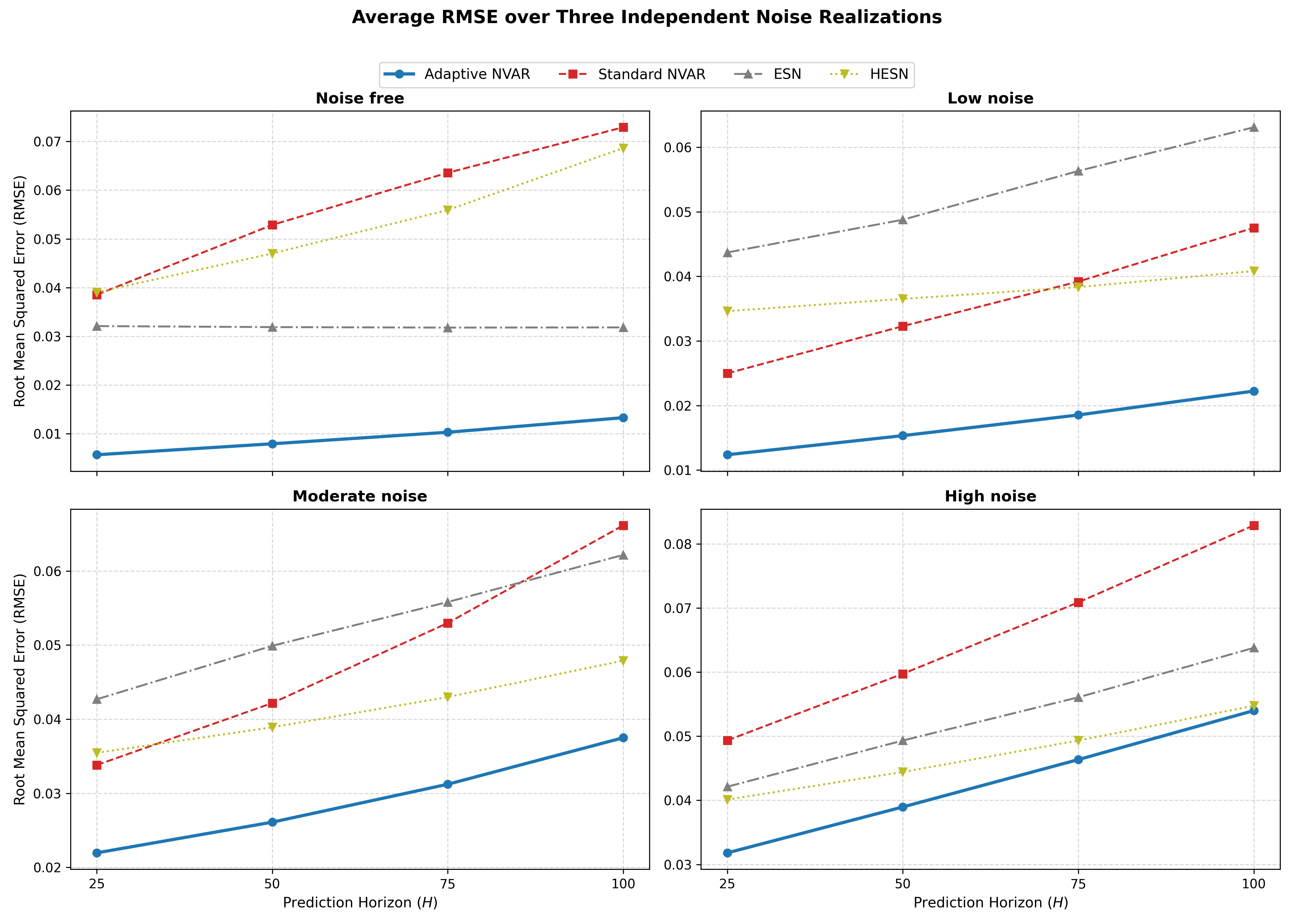}
    \caption{\textbf{Average forecasting performance across noise levels for the Mackey--Glass system.} 
    Results show the average RMSE over the three independent noise realizations presented in Fig.~\ref{fig:mg_results}.}
    \label{fig:mg_summary}
\end{figure}

To summarize the overall trends, Fig.~\ref{fig:mg_summary} reports the average RMSE over the three independent noise realizations (Datasets 1--3) for each noise level. Averaging over multiple realizations reduces the influence of individual runs, providing a more reliable assessment of each model's robustness.

The averaged results confirm the observations from Fig.~\ref{fig:mg_results}. Adaptive NVAR consistently produces the lowest RMSE under all noise conditions and across all prediction horizons. Furthermore, its prediction error increases more slowly with forecasting horizon than those of the benchmark models, demonstrating superior long-term recursive forecasting stability.

The benchmark models exhibit distinct behaviors as the noise increases. Under noise-free conditions, ESN provides the second-best performance after Adaptive NVAR. However, as the observation noise increases, HESN consistently outperforms ESN, indicating that incorporating knowledge-based reservoir dynamics improves robustness to noisy observations. Nevertheless, both reservoir-based models remain consistently less accurate than Adaptive NVAR across all experimental settings. The complete numerical results corresponding to Figs.~\ref{fig:mg_results} and \ref{fig:mg_summary} are provided in Supplementary Table~4 for reproducibility and quantitative comparison.

\subsection{Low-Dimensional System: Lorenz–63}

The second chaotic system considered is the Lorenz--63 system \cite{lorenz2017deterministic}, which is a three-dimensional chaotic dynamical system governed by the equations
\begin{equation*}
\begin{aligned}
\frac{dx}{dt}(t) &= \sigma \bigl(y(t) - x(t)\bigr), \\
\frac{dy}{dt}(t) &= x(t)\,\bigl(\rho - z(t)\bigr) - y(t), \\
\frac{dz}{dt}(t) &= x(t)\,y(t) - \beta\,z(t).
\end{aligned}
\end{equation*}
We adopt the parameter values $\sigma = 10,\; \rho = 28$, and $\beta = \tfrac{8}{3}$, with initial conditions 
\[
(x(0),y(0),z(0)) = (-8.0,\,7.0,\,27.0).
\]

\begin{figure}[h!]
    \centering
    \includegraphics[width=\linewidth]{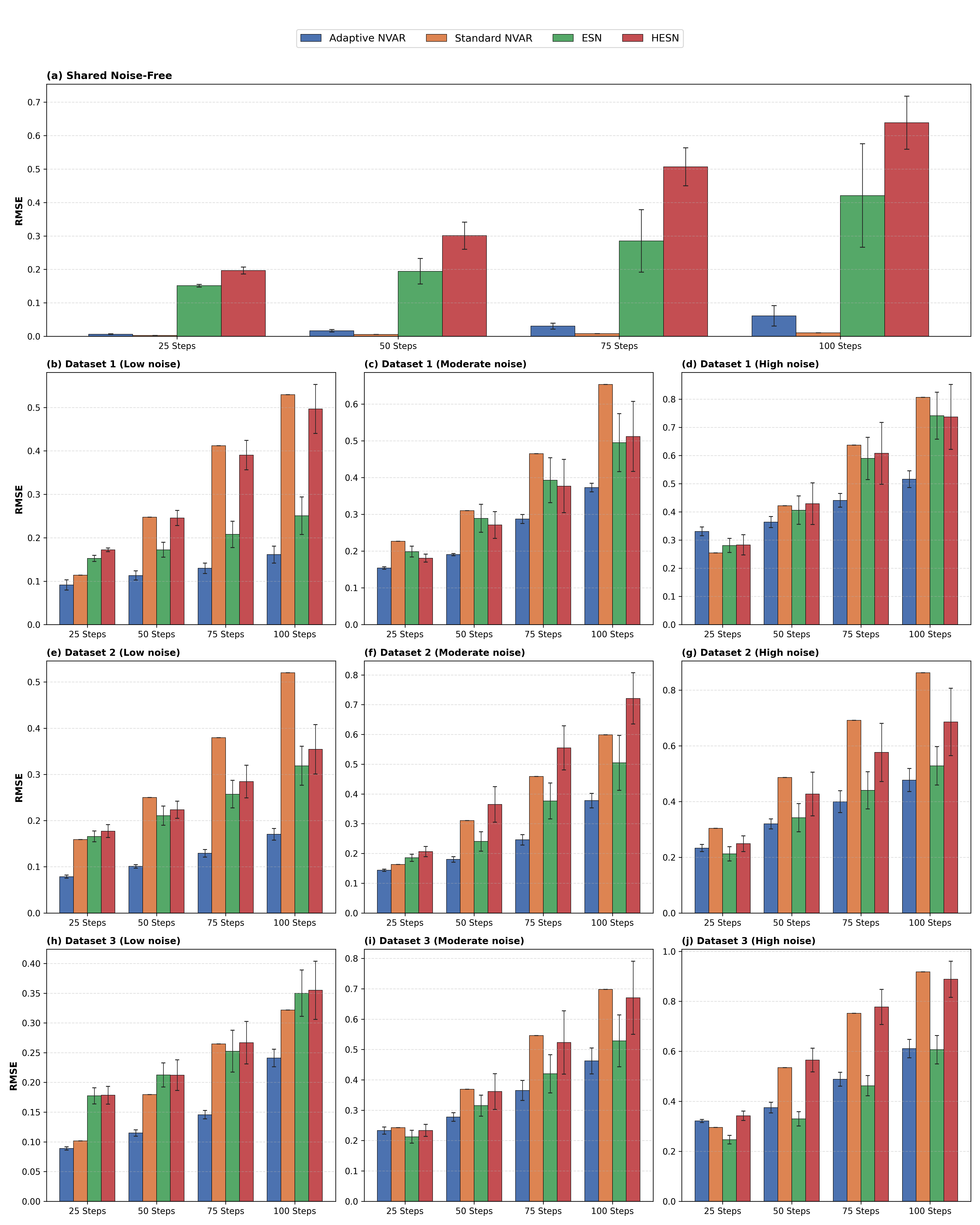}
    \caption{\textbf{Predictive performance and robustness to noise for the Lorenz 63 system.} 
     (a) Noise-free forecasting performance. (b--j) Forecasting performance under three independent noise realizations (Datasets~1--3) at low ($\sigma=0.10$), moderate ($\sigma=0.20$), and high ($\sigma=0.30$) observation noise levels. Bar heights denote the mean RMSE over overlapping forecast windows, while error bars indicate the standard deviation. Forecasting performance is evaluated at prediction horizons of 25, 50, 75, and 100 time steps}
    \label{fig:l63_results}
\end{figure}

\begin{figure}[h!]
    \centering
    \includegraphics[width=\linewidth]{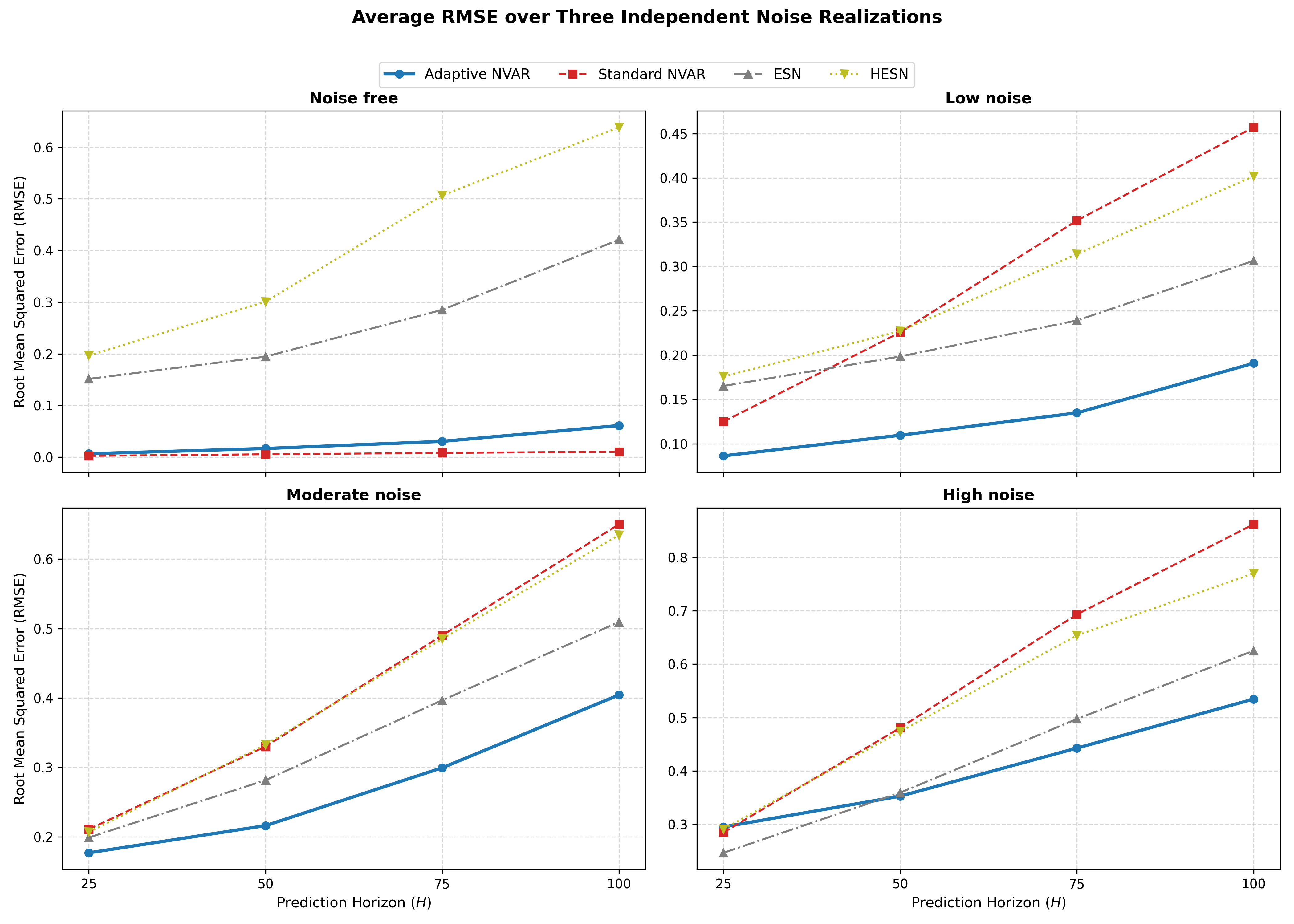}
    \caption{\textbf{Average forecasting performance across noise levels for the Lorenz 63 system.} 
    Results show the average RMSE over the three independent noise realizations presented in Fig.~\ref{fig:l63_results}.}
    \label{fig:l63_summary}
\end{figure}

The system was numerically integrated using a Runge–Kutta RK45 solver with a time step of $\Delta t = 0.001$. The data were sampled every $20\Delta t$ to produce a time series of $5{,}000$ data points. An $80/10/10$ split was applied for warm-up $+$ training/validation$/$testing, respectively. Specifically, the first $100$ points were discarded as warm-up for all models. The remaining data were used for training, followed by $500$ points for validation and the final $500$ for testing. For the HESN model, a knowledge-based model is produced by substituting $\beta$ for $(1 + \epsilon)\beta$, with an error parameter of $\epsilon = 0.05$.

Figure~\ref{fig:l63_results} presents the detailed forecasting performance of the proposed Adaptive NVAR model and the three benchmark models under noise-free and noisy scenarios. The noise-free results are shown in Fig.~\ref{fig:mg_results}(a), while Figs.~\ref{fig:l63_results}(b–j) report the results for three independent noise realizations (Datasets 1--3) under low, moderate, and high noise levels.

Under noise-free conditions, NG-RC achieves the lowest RMSE across all prediction horizons, while Adaptive NVAR consistently provides the second-best performance and substantially outperforms both ESN and HESN. At the longest prediction horizon of 100 steps, NG-RC attains an RMSE of 0.0102, compared with 0.0610 for Adaptive NVAR, 0.4207 for ESN, and 0.6384 for HESN. Although Adaptive NVAR exhibits slightly larger errors than NG-RC under ideal observation conditions, the difference remains relatively small compared with the considerably larger errors of the reservoir-based approaches.

An interesting observation is that NG-RC performs considerably better on the Lorenz-63 benchmark than on the Mackey-Glass system under noise-free conditions. This behavior can be attributed to both the optimization strategy and the characteristics of the underlying dynamics. NG-RC computes the readout weights using the closed-form solution of ridge regression over a fixed polynomial features, yielding the optimal linear readout for the selected features. Moreover, the Lorenz-63 system is governed primarily by low-order polynomial interactions, which closely match the predefined polynomial basis employed by NG-RC. In contrast, the proposed Adaptive NVAR jointly learns adaptive feature-selection parameters through gradient-based optimization, resulting in a non-convex optimization problem whose solution is generally locally optimal. Consequently, a slight reduction in prediction accuracy under ideal noise-free conditions is expected. Nevertheless, Adaptive NVAR remains highly competitive while offering significantly improved robustness once measurement noise is introduced.

The robustness of the models under noise is illustrated in Figs.~\ref{fig:l63_results}(b–j). As expected, the prediction error increases with both forecasting horizon and noise for all methods. However, unlike the noise-free case, Adaptive NVAR consistently achieves the lowest RMSE across nearly all prediction horizons, all three independent noise realizations, and all noise levels. The performance advantage becomes increasingly pronounced as the noise increases, demonstrating that the adaptive feature-selection mechanism effectively suppresses the influence of noisy measurements while preserving the underlying system dynamics.

In contrast, NG-RC experiences a substantial degradation in forecasting accuracy as the noise increases, despite its excellent performance under noise-free conditions. ESN and HESN also exhibit larger prediction errors throughout the noisy experiments, with HESN generally providing improved robustness over ESN but remaining consistently less accurate than Adaptive NVAR. Furthermore, Adaptive NVAR maintains relatively small standard deviations across multiple runs, indicating stable optimization and reliable forecasting performance under noisy cases.

To summarize the overall trends, Fig.~\ref{fig:l63_summary} reports the average RMSE over the three independent noise realizations (Dataset 1--3) for each noise level. The averaged results confirm the observations from Fig.~\ref{fig:l63_results}. Although NG-RC remains the most accurate model under noise-free conditions, Adaptive NVAR consistently achieves the lowest average RMSE under low, moderate, and high noise conditions across all prediction horizons. Moreover, its prediction error increases more gradually with forecasting horizon than those of the benchmark models, demonstrating superior long-term recursive forecasting stability in the presence of noise. 

The complete numerical results corresponding to Figs.~\ref{fig:l63_results} and \ref{fig:l63_summary} are provided in Supplementary Table~5.

\subsection{High-Dimensional System: Lorenz–96}\label{Sec:Lorenz96}

The final system evaluated to demonstrate scalability is the Lorenz--96 model, which was first introduced by Lorenz \cite{lorenz1996predictability} and is governed by the following system of coupled differential equations:
\begin{equation*}
\frac{dx_i}{dt} = \bigl(x_{i+1} - x_{i-2}\bigr)\,x_{i-1} - x_i + F, 
\qquad i=1,\dots,N,
\end{equation*}
subject to the cyclic boundary conditions:
\[
x_{i-N}=x_{i+N} = x_i.
\]
Owing to its circulant symmetry, this model is frequently used as a simplified depiction of atmospheric fluid dynamics. Here, the index $i$ represents discrete longitudinal points along a constant circle of latitude, and the state variable $x_i$ denotes a localized atmospheric quantity. 

To evaluate the models under highly chaotic regimes, experiments were conducted with a spatial dimension of $N = 100$ variables and an external forcing parameter of $F = 8$. A fourth-order Runge--Kutta (RK4) solver with an integration step size of $\Delta t = 0.001$ was used to integrate the system numerically. The resulting trajectory was downsampled by a factor of 10 to generate a dataset spanning $500{,}000$ temporal points. An $80/10/10$ split was implemented for the training, validation, and testing partitions, respectively, with the initial $1{,}000$ steps discarded as a washout (warm-up) phase. 

As detailed in Sec.~\ref{Sec:scal}, the minimum number of training instances required by the standard NVAR framework is strictly bounded by the size of its nonlinear feature vector. Consequently, while a low-dimensional system like Lorenz-63 requires roughly $4{,}000$ training steps, scaling to a Lorenz-96 system with $N=100$ variables requires at least around $500{,}000$ points to prevent the underlying linear system of equations from becoming underdetermined. This benchmark therefore serves as an excellent testbed for evaluating the scalability limits of different forecasting methodologies under high-dimensional chaos.

For a fair evaluation across systems, initial parameters were structured around $m$. However, the projection layer dimension $d_{\mathcal{NN},i}$ is treated as a tunable hyperparameter that does not need to equal the full quadratic expansion dimension. To achieve computational efficiency, we define:
\begin{equation}\label{cdk}
d_{\mathcal{NN},i} = c \cdot d \cdot k,
\end{equation}

where $c$ is a small constant scaling factor that regulates the retention of nonlinear state information. This formulation preserves the necessary nonlinear representational capacity while effectively mitigating the dimensionality explosion typical of fixed polynomial expansions. The scaling constant $c$ was treated as a tunable parameter optimized via TPE. Validating across a search space of $c \in \{2, 5, 10\}$, the TPE identified $c=2$ as the optimal configuration for all noisy scenarios, whereas $c=5$ yielded the best prediction under a noise-free scenario.
Remarkably, only the proposed Adaptive NVAR framework successfully completed both the training and forecasting stages within the constraints of our computational environment (see Sec.~\ref{Sec:computational-environment} for hardware details). In our specific experimental setup ($d=100$, $k=10$, and $c=2$), the output dimension of the multilayer perceptron (MLP) yields $d_{\mathcal{NN},i} = 2 \times 100 \times 10 = 2{,}000$ nodes. Notably, even with a conservative choice of $c=2$, the resulting feature space dimension $d_{\mathcal{NN},i}$ remains approximately 250 times smaller than the full quadratic feature matrix size ($m \approx 5 \times 10^5$). This structural reduction drops the overall network footprint down to a few megabytes, enabling fast GPU-accelerated backpropagation. Moreover, Adaptive NVAR was trained using mini-batch gradient optimization with a batch size of 1024. This avoids constructing the full training feature matrix in memory and enables efficient GPU-accelerated optimization on large-scale datasets

In contrast, severe memory bottlenecks completely inhibited standard NVAR from scaling to the 100-variable Lorenz–96 system. Because the standard NVAR framework computes its weight parameters via a closed-form ridge regression, the combinatorial growth of its feature vector requires the direct inversion of an exceptionally large matrix, causing instant system out-of-memory errors (see Sec.~\ref{Sec:scal}). Similarly, executing a rigorous hyperparameter grid search or TPE optimization for traditional reservoir-based models (ESN and HESN) at $d=100$ proved computationally intractable due to the immense number of internal parameters (e.g., reservoir size, spectral radius) requiring simultaneous tuning. While alternative literature has addressed reservoir scalability via local-state architectures \cite{parlitz2000prediction, baur2021predicting}, the objective of this work is specifically focused on addressing the inherent limitations of standard NVAR while validating performance under strictly controlled baseline conditions. By learning nonlinear features through a shared MLP rather than explicitly constructing a polynomial feature matrix, Adaptive NVAR can naturally be trained using mini-batch optimization, making it more suitable for large-scale datasets.

\begin{figure}[h!]
    \centering
    \includegraphics[width=0.9\linewidth]{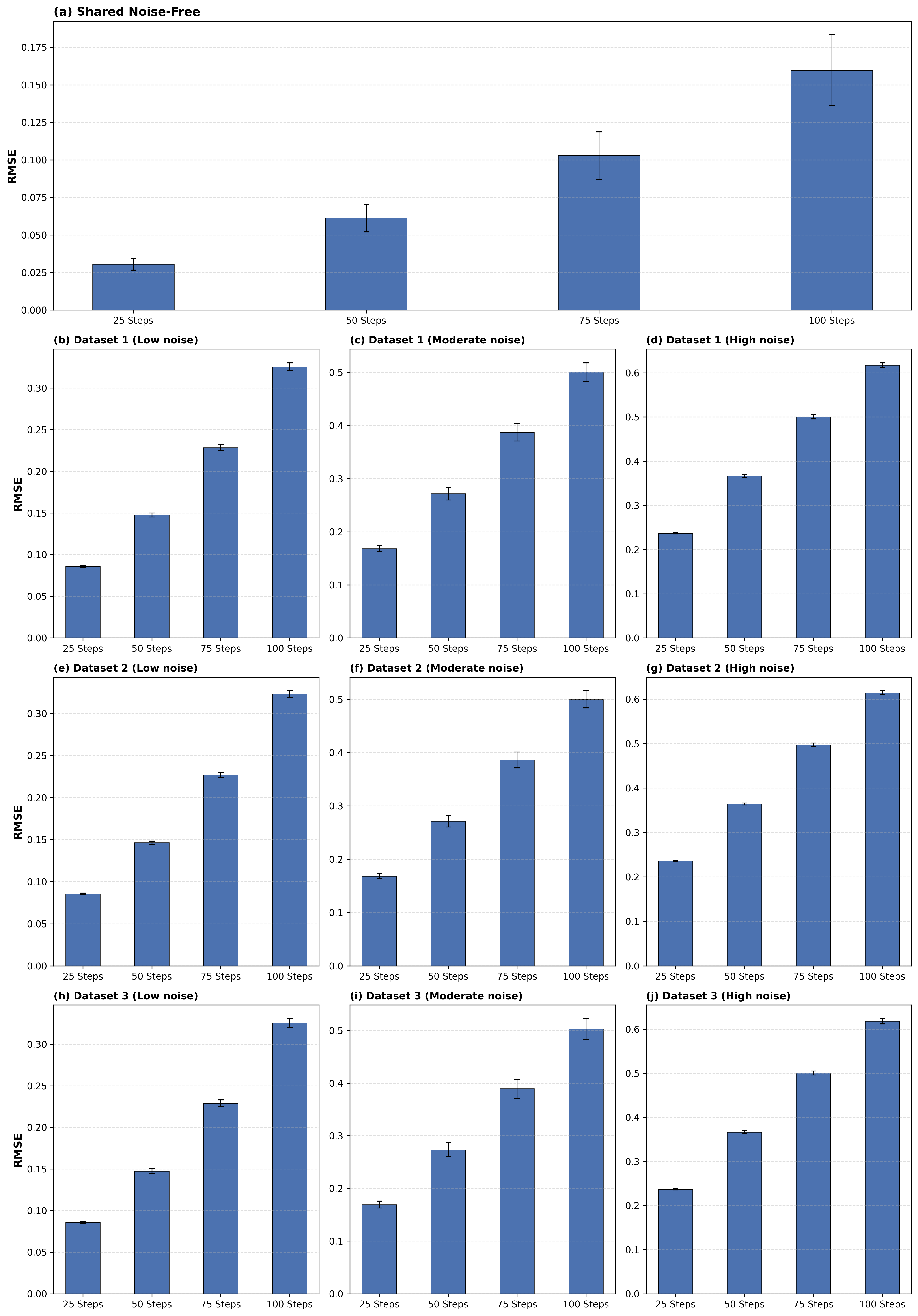}
    \caption{\textbf{Predictive performance and robustness to  noise   for the Lorenz 96 ($N=100$) system.} 
     (a) Noise-free forecasting performance. (b--j) Forecasting performance under three independent noise realizations (Datasets~1--3) at low ($\sigma=0.10$), moderate ($\sigma=0.20$), and high ($\sigma=0.30$) noise levels. Bar heights denote the mean RMSE over overlapping forecast windows, while error bars indicate the standard deviation. Forecasting performance is evaluated at prediction horizons of 25, 50, 75, and 100 time steps}
    \label{fig:l96_results}
\end{figure}

\begin{figure}[h!]
    \centering
    \includegraphics[width=\linewidth]{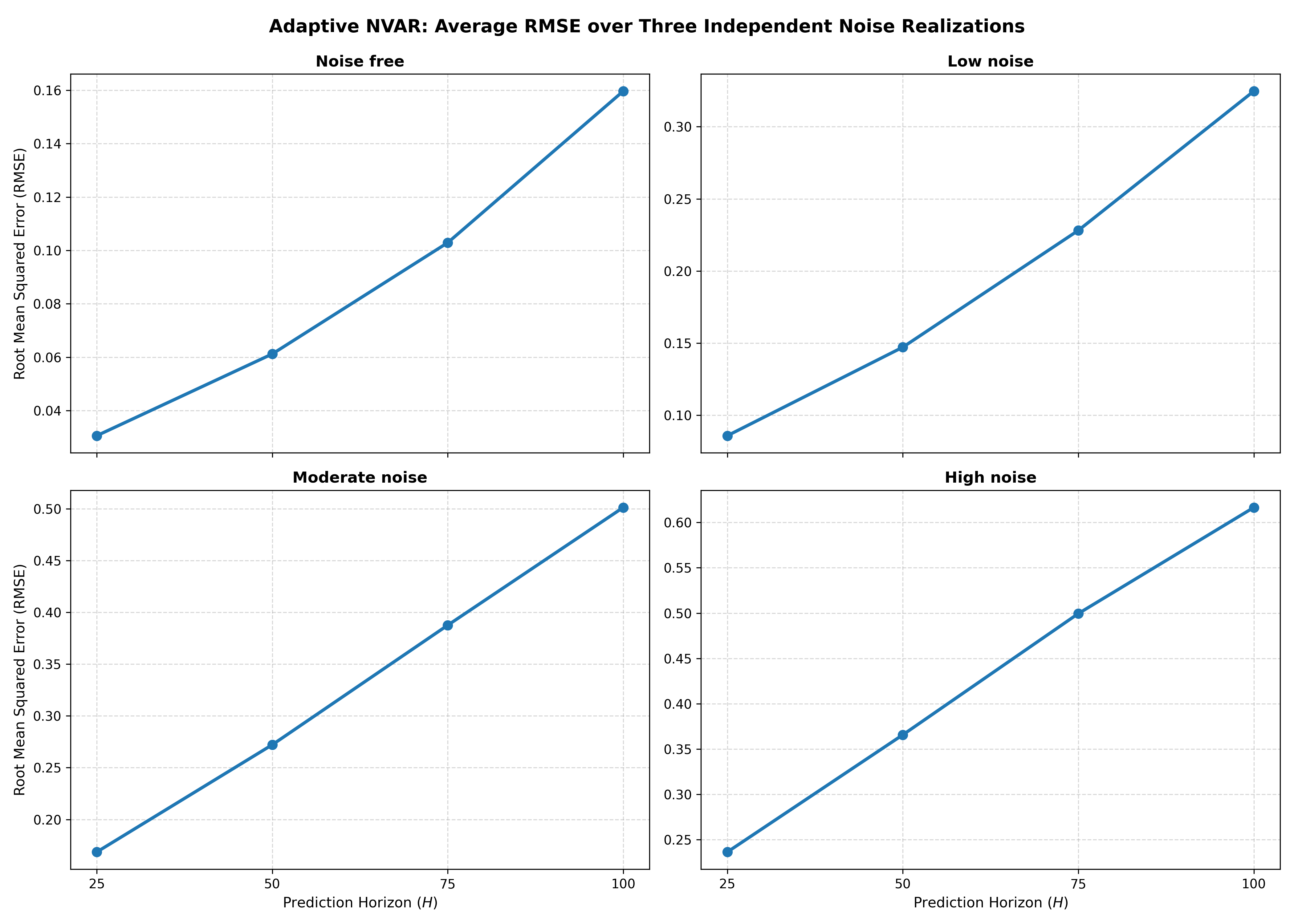}
    \caption{\textbf{Average forecasting performance across noise levels for the Lorenz 96 ($N=100$) system.} 
    Results show the average RMSE over the three independent noise realizations presented in Fig.~\ref{fig:l96_results}.}
    \label{fig:l96_summary}
\end{figure}

Detailed multi-horizon forecasting performance of Adaptive NVAR under different noise conditions is illustrated in Figures~\ref{fig:l96_results} and \ref{fig:l96_summary}. Figure~\ref{fig:l96_results} presents the forecasting accuracy across the considered prediction horizons. Under the noise-free case (Fig.~\ref{fig:l96_results}(a)), Adaptive NVAR accurately tracks the high-dimensional attractor, achieving low mean test RMSE values that increase gradually from $0.0306$ at $H=25$ to $0.1597$ at $H=100$. 

Under noisy observations (Figs.~\ref{fig:l96_results}(b)--(j)), Adaptive NVAR maintains consistent forecasting performance across all three independent noise realizations (Datasets 1--3), demonstrating strong robustness to measurement noise. For a given noise level, the error profiles are highly consistent across datasets. Under low noise ($\sigma=0.10$), the RMSE at the 100-step horizon remains around $0.325$, whereas under high noise ($\sigma=0.30$), it increases consistently to approximately $0.617$. 
To highlight the overall error growth, Fig.~\ref{fig:l96_summary} presents the mean RMSE averaged over the three independent noise realizations for each noise level. The averaged curves show a smooth, nearly linear increase in error with prediction horizon, indicating stable degradation of forecasting accuracy as the horizon extends rather than abrupt error growth. The complete numerical results corresponding to Figs.~\ref{fig:l96_results} and \ref{fig:l96_summary} are provided in Supplementary Table~6 for reproducibility and quantitative comparison.

\section{Discussion}\label{Sec:discussion}
\subsection{Structural Mechanics and Mathematical Advantages under Noise}

The experiments on the Mackey--Glass, Lorenz 63, and Lorenz 96 systems demonstrate that Adaptive NVAR consistently provides improved robustness to noise while maintaining competitive forecasting accuracy in the absence of noise. This behavior can be understood from the architecture of the proposed model and the mathematical properties of its learned nonlinear feature representation.

The principal architectural distinction of Adaptive NVAR is its parallel skip-connected feature representation. The original delay embedding is passed directly to the final linear readout while, in parallel, a shallow MLP learns complementary nonlinear features from the same embedding. If the learned nonlinear features provide little additional predictive value, the readout can assign them negligible weights, effectively reducing the model to a linear delay-coordinate predictor. Consequently, the proposed architecture increases the model's representational capacity without discarding the information contained in the original delay embedding.

By contrast, standard NVAR constructs its nonlinear representation through a fixed quadratic polynomial expansion. Suppose the observed state is contaminated by additive measurement noise,
\begin{equation}
\tilde{\mathbf X}(t)
=
\mathbf{X_\mathrm{clean}}(t)
+
\epsilon(t),
\qquad
\epsilon(t)\sim\mathcal N(0,\sigma^2).
\end{equation}
The quadratic feature expansion then produces interaction terms of the form
\begin{equation}
(x_i+\epsilon_i)(x_j+\epsilon_j)
=
x_ix_j
+
x_i\epsilon_j
+
x_j\epsilon_i
+
\epsilon_i\epsilon_j.
\end{equation}
These additional terms introduce state-dependent interactions between the signal and the noise, together with higher-order noise terms. Since the polynomial feature is fixed \emph{a priori}, the subsequent ridge regression can only determine the optimal linear readout within this predetermined feature space; it cannot modify the nonlinear representation itself. Consequently, if the predefined features are poorly suited to noisy observations, the model has no mechanism for adapting its feature representation.

Adaptive NVAR addresses this limitation by replacing the handcrafted polynomial expansion with a trainable nonlinear mapping. The adaptive branch consists of a shallow multilayer perceptron in Eq.~\ref{eq:H_NN} with a $\tanh$ activation function whose derivative satisfies
\begin{equation}
\frac{d}{dz}\tanh(z)
=
1-\tanh^2(z)
\le
1.
\end{equation}
Therefore, the hyperbolic tangent is a $1$-Lipschitz activation function. Since affine transformations are linear operators with finite induced norms, the neural feature mapping is also Lipschitz continuous. Specifically, for a perturbation $\epsilon$ in the delay-embedded input,
\begin{equation}
\begin{aligned}
&
\left\|
H_{\mathcal{NN}}(H_{\mathrm{lin}}+\epsilon)
-
H_{\mathcal{NN}}(H_{\mathrm{lin}})
\right\|
\\
&\le
\|W\|
\,
\left\|
\tanh\!\left(
W_{\mathrm{in}}(H_{\mathrm{lin}}+\epsilon)+b_1
\right)
-
\tanh\!\left(
W_{\mathrm{in}}H_{\mathrm{lin}}+b_1
\right)
\right\|
\\
&\le
\|W\|
\,
\|W_{\mathrm{in}}\|
\,
\|\epsilon\|,
\end{aligned}
\end{equation}
where $\|\cdot\|$ denotes the induced operator norm.

This bound shows that perturbations in the learned nonlinear representation grow at most linearly with the magnitude of the input disturbance. Unlike the fixed quadratic expansion, whose feature construction explicitly introduces higher-order interactions involving the noise, the learned nonlinear mapping remains uniformly Lipschitz continuous with respect to its input. Moreover, the bounded $\tanh$ activation limits the magnitude of its outputs, reducing the sensitivity of the learned representation to large input perturbations.

Finally, the nonlinear feature mapping and the linear readout are optimized jointly through end-to-end minimization of the forecasting loss. Rather than relying on a predetermined nonlinear basis, Adaptive NVAR learns a task-specific feature representation that is optimized directly for prediction. Combined with the preserved linear delay embedding provided by the skip connection, this adaptive feature learning offers a plausible explanation for the improved robustness observed across the noisy Mackey--Glass, Lorenz 63, and Lorenz 96 forecasting experiments.

\subsection{System-Specific Behaviors: Lorenz 63 vs. Mackey--Glass}
The distinct behavior of standard NG-RC between the Lorenz '63 and Mackey--Glass benchmarks validates our theoretical rationale regarding fixed versus adaptive features. 

Under noise-free conditions, standard NG-RC performs exceptionally well on the Lorenz-63 system, achieving a Horizon 100 RMSE of $0.0102$. This is expected because the Lorenz-63 dynamics are governed by low-order quadratic interactions (e.g., $xy$ and $xz$), which are well captured by the predefined quadratic feature features used in standard NVAR. As a result, the closed-form ridge regression can accurately determine the readout weights, leading to highly accurate predictions. However, once  even the low noise is introduced, the explicit polynomial feature expansion also amplifies the noise, causing the Horizon 100 RMSE to increase sharply to $0.9183$.

The Mackey--Glass system presents a different challenge. As a nonlinear delay differential equation with an infinite-dimensional phase space and long-range temporal dependencies, its dynamics are much more difficult to capture using a fixed quadratic feature expansion. Consequently, reservoir-based methods such as ESN and HESN become more prone to trajectory divergence under noisy conditions. In addition, the fixed structural mismatch incorporated into HESN introduces a persistent modeling bias that further limits its predictive accuracy. These observations underscore the importance of learning flexible nonlinear representations when forecasting complex delay systems in the presence of noise.

\subsection{Computational Cost and Scalability of NVAR Frameworks}\label{Sec:scal}

A fundamental limitation of the standard NVAR framework is the rapid growth of its explicit polynomial feature space as the system dimension increases. As described in Section~\ref{Sec:NVAR}, the dimensionality of the quadratic feature vector is
\begin{equation}
d_{\text{total}} = 1 + d_{\text{lin}} + d_{\text{nonlin}}, \qquad
d_{\text{lin}} = dk, \qquad
m := d_{\text{nonlin}} = \frac{d_{\text{lin}}(d_{\text{lin}}+1)}{2},
\end{equation}
which scales quadratically with both the system dimension $d$ and the delay length $k$. Consequently, increasing either parameter rapidly enlarges the feature space, substantially increasing both memory requirements and computational cost.
Given the full feature matrix $\mathbf{H}_{\text{total}} \in \mathbb{R}^{d_{\text{total}} \times T}$, where each column corresponds to one time sample of the full NVAR feature vector across $T$ total training points, the analytical ridge regression solution requires computing the inverse of the Gram matrix: \begin{equation} \mathbf{H}_{\text{total}} \mathbf{H}_{\text{total}}^{\top} \in \mathbb{R}^{d_{\text{total}} \times d_{\text{total}}} \end{equation} The rank of this Gram matrix satisfies the algebraic condition: \begin{equation} \operatorname{rank}\!\left(\mathbf{H}_{\text{total}} \mathbf{H}_{\text{total}}^{\top}\right) \le \min\left(d_{\text{total}}, T\right) \end{equation} Consequently, if $d_{\text{total}} > T$, the unregularized Gram matrix is strictly rank-deficient. While adding a ridge regularization penalty $\alpha \mathbf{I}$ renders the system analytically invertible, the optimization problem remains fundamentally underdetermined. It is mathematically impossible to robustly resolve $d_{\text{total}}$ independent feature mappings from fewer than $d_{\text{total}}$ unique data observations, leaving the regression highly sensitive to noise and heavily biased by the choice of $\alpha$. Reliable execution therefore demands a data regime where $T \gg d_{\text{total}}$.

The proposed Adaptive NVAR addresses this limitation by replacing the explicit polynomial feature expansion with a shared learned nonlinear representation. Rather than constructing the complete quadratic feature, a shallow MLP learns a shared nonlinear mapping from the delay embedding, while the original linear embedding is preserved through a skip connection. This allows the dimension of the learned nonlinear representation to be treated as a tunable hyperparameter instead of being fixed by the combinatorial size of the polynomial basis.

An additional advantage of this formulation is its compatibility with gradient-based optimization. Unlike the closed-form ridge regression used in standard NVAR, Adaptive NVAR can be trained using mini-batch optimization, enabling large datasets to be processed sequentially without explicitly constructing the complete polynomial feature matrix in memory. As demonstrated in the Lorenz--96 experiments, this significantly improves the practicality of applying NVAR-based forecasting to high-dimensional chaotic systems.

\subsection{Architectural Taxonomy}
To contextualize these distinct paradigms, Table~\ref{tab:nvar_esn_summary} provides a comprehensive conceptual and computational taxonomy characterizing standard NVAR, Echo State Networks variants, and our proposed Adaptive NVAR framework.

\begin{table}[h!]
\centering
\caption{Conceptual and computational comparison of NVAR variants and ESN/HESN models.}
\label{tab:nvar_esn_summary}
\renewcommand{\arraystretch}{1.25}
\setlength{\tabcolsep}{5pt}
\small
\begin{tabular}{|p{3.5cm}|p{3.2cm}|p{3.2cm}|p{3.2cm}|}
\hline
\textbf{Property} & \textbf{Standard NVAR} & \textbf{Adaptive NVAR} & \textbf{ESN / HESN} \\
\hline
\textbf{Feature construction} & Delay embedding + polynomial expansion & Delay embedding + trainable shallow MLP & Random or hybrid reservoir dynamics generate nonlinear features \\
\hline
\textbf{Randomness} & None & From initial MLP weights (trainable) & Strong dependence on random reservoir; HESN adds structured knowledge input \\
\hline
\textbf{Training} & Closed-form ridge regression (convex) & Gradient-based backpropagation with Adam  & Closed-form ridge regression (convex) \\
\hline
\textbf{Noise robustness} & Sensitive (polynomial terms amplify noise) & High (adaptive mapping enables denoising) & Mixed (reservoir dynamics may amplify or dampen noise) \\
\hline
\textbf{Computational cost / scalability} & High cost, poor scaling with input dimension & Moderate cost, scalable via shallow MLP mini-batches & High cost, limited by reservoir size and structural tuning gridlock \\
\hline
\textbf{Expressivity} & Limited (fixed polynomial basis) & High (learned nonlinear mapping) & High (nonlinear reservoir dynamics) \\
\hline
\end{tabular}
\end{table}
\subsection{Limitations and Future Directions}

Although the proposed Adaptive NVAR framework demonstrates strong forecasting performance and robustness under additive Gaussian measurement noise, this study considers only one type of observational uncertainty. In practice, real-world time-series data are often affected by more complex disturbances, including non-Gaussian and temporally correlated noise, sensor drift, missing observations, outliers, and measurement artifacts.

Because Adaptive NVAR learns its nonlinear feature representation directly from data, it may also be effective under more challenging noise conditions. However, this capability has not been systematically evaluated in the present work. Therefore, the results should be interpreted as demonstrating robustness specifically to Gaussian measurement noise rather than to all forms of data corruption.

The proposed framework also employs a single shallow MLP across all experiments. Despite its simplicity, this architecture consistently improves forecasting accuracy and noise robustness, suggesting that even a lightweight learned nonlinear representation can be highly effective. Nevertheless, exploring deeper or alternative network architectures may further improve performance, particularly for more complex dynamical systems. In addition, a direct quantitative comparison with standard NVAR on the Lorenz 96 benchmark was not feasible because the quadratic polynomial feature expansion exceeded the available computational resources.

Future work will evaluate Adaptive NVAR under a wider range of realistic noise models and investigate alternative neural architectures for nonlinear feature learning. We also plan to validate the proposed framework on real-world chaotic forecasting problems, including high-resolution sea surface temperature prediction.

\subsection{Computational Environment}\label{Sec:computational-environment}
All experiments and analyses were carried out in Jupyter Notebook~\cite{JN}, which integrates code execution, visualization, and narrative documentation in a single environment. All computations were performed on the system configuration summarized in Table~\ref{tab:environment}. The main libraries used included Matplotlib~\cite{plt} for visualization, NumPy~\cite{np} and SciPy~\cite{Sci} for numerical computation, and PyTorch~\cite{PyTorch} for model implementation.

\begin{table}[h!]
\centering
\caption{Runtime environment used for the experiments.}
\label{tab:runtime_environment}
\renewcommand{\arraystretch}{1.25}

\begin{tabular}{lc}
\toprule
\textbf{Name} & \textbf{Configuration} \\
\midrule
CPU
& Intel(R) Xeon(R) Gold 6242 CPU @ 2.80GHz \\
GPU
& NVIDIA A100-PCIE-40GB \\
Memory
& 376.54 GB RAM \\
Operating System
& Linux 3.10.0-1062.el7.x86\_64 (64-bit) \\
\bottomrule
\end{tabular}
\label{tab:environment}
\end{table}

\section*{Data Availability}\label{Sec:data}
Given the sensitivity of chaotic systems to numerical precision and hardware-specific variations, we make available the full simulation dataset together with the source code, thereby ensuring reproducibility of all experiments. All data and accompanying code are available at \href{https://github.com/AzimovSherkhon/Adaptive-Nonlinear-Vector-Autoregression-Robust-Forecasting-for-Noisy-Chaotic-Time-Series.git}{github.com/AzimovSherkhon/Adaptive-Nonlinear-Vector-Autoregression-Robust-Forecasting-for-Noisy-Chaotic-Time-Series.git}.

\section*{Code Availability}\label{Sec:code}

To promote transparency and reproducibility, the source code implementing Adaptive NVAR has been made publicly available at \href{https://github.com/AzimovSherkhon/Adaptive-Nonlinear-Vector-Autoregression-Robust-Forecasting-for-Noisy-Chaotic-Time-Series.git}{github.com/AzimovSherkhon/Adaptive-Nonlinear-Vector-Autoregression-Robust-Forecasting-for-Noisy-Chaotic-Time-Series.git}.

\section*{Acknowledgments}
This work was supported by a National Research Foundation of Korea (NRF) grant funded by the Korea Government (MSIT) (2022R1A5A1033624; RS-2023-00242528); the Korea Institute of Marine Science \& Technology Promotion (KIMST), funded by the Ministry of Oceans and Fisheries (RS-2025-02217872); and the Global- Learning \& Academic research institution for Master’s · Ph.D. students, and Postdocs (LAMP) Program of the National Research Foundation of Korea (NRF) grant, funded by the Ministry of Education (No. RS-2023- 00301938).

Additionally, the work of S. López-Moreno was supported by the Korea National Research Foundation (NRF) grant funded by the Korean government (MSIT) (RS-2024-00406152), and the work of E. Dolores Cuenca was supported by the Korea National Research Foundation (NRF) grant funded by the Korean government (MSIT) (RS-2025-00517727).

\section*{Author Contributions}
AS: Conceptualization, Data curation, Investigation, Methodology, Software, Validation, Visualization, Writing – original draft; SL-M: Investigation, Validation, Visualization, Writing – review and editing; ED-C: Formal Analysis, Investigation, Validation, Writing – review and editing; SL: Software, Visualization; JK: Funding acquisition, Computational resources, Project administration; SK: Funding acquisition, Project administration, Supervision, Writing – review and editing.

\section*{Competing Interests}
The authors declare no competing interests.

\section*{Additional Information}
\subsection*{Supplementary information}
Supplementary material available at \href{https://github.com/AzimovSherkhon/Adaptive-Nonlinear-Vector-Autoregression-Robust-Forecasting-for-Noisy-Chaotic-Time-Series.git}{github.com/AzimovSherkhon/Adaptive-Nonlinear-Vector-Autoregression-Robust-Forecasting-for-Noisy-Chaotic-Time-Series.git}

\subsection*{Correspondence}
Correspondence and requests for materials should be addressed to Sangil Kim (email: sangil.kim@pusan.ac.kr) and Sherkhon Azimov (email: sherxonazimov94@pusan.ac.kr).

\printbibliography


\end{document}